\documentclass[conference]{IEEEtran}
\IEEEoverridecommandlockouts
\usepackage{cite}
\usepackage{url}
\usepackage{amsmath,amssymb,amsfonts}
\usepackage{algorithmic}
\usepackage{float}
\usepackage[most]{tcolorbox}
\usepackage{graphicx}
\usepackage{adjustbox}
\graphicspath{ {./figures/} } 
\usepackage{textcomp}
\usepackage{hyperref}
\usepackage{tikz}
\usetikzlibrary{shapes.multipart, positioning, arrows.meta, fit, calc}
\hypersetup{
    colorlinks=false,
    urlcolor=blue,
}
\def\BibTeX{{\rm B\kern-.05em{\sc i\kern-.025em b}\kern-.08em
    T\kern-.1667em\lower.7ex\hbox{E}\kern-.125emX}}
\begin{document}

\title{An Agentic Flow for Finite State Machine Extraction using Prompt Chaining \\
{\footnotesize \textsuperscript{}}}

\author{\IEEEauthorblockN{1\textsuperscript{st} Fares Wael}
\IEEEauthorblockA{\textit{Dept. of Computer Science} \\
\textit{MSA University}\\
Giza, Egypt \\
fares.wael@msa.edu.eg}

\and
\IEEEauthorblockN{2\textsuperscript{nd} Youssef Maklad}
\IEEEauthorblockA{\textit{Dept. of Computer Science} \\
\textit{MSA University}\\
Giza, Egypt \\
youssef.mohamed88@msa.edu.eg}

\and
\IEEEauthorblockN{3\textsuperscript{rd} Ali Hamdi}
\IEEEauthorblockA{\textit{Dept. of Computer Science} \\
\textit{MSA University}\\
Giza, Egypt \\
ahamdi@msa.edu.eg}

\and
\IEEEauthorblockN{4\textsuperscript{th} Wael Elsersy}
\IEEEauthorblockA{\textit{Dept. of Computer Science} \\
\textit{MSA University}\\
Giza, Egypt \\
wfarouk@msa.edu.eg}
}


\IEEEoverridecommandlockouts
\IEEEpubid{\makebox[\columnwidth]{ 979-8-3315-0185-3/25/\$31.00 ©2025 IEEE \hfill}
\hspace{\columnsep}\makebox[\columnwidth]{ }}
\maketitle
\IEEEpubidadjcol

\begin{abstract}
Finite-State Machines (FSMs) are critical for modeling the operational logic of network protocols, enabling verification, analysis, and vulnerability discovery. However, existing FSM extraction techniques face limitations such as scalability, incomplete coverage, and ambiguity in natural language specifications. In this paper, we propose \textit{FlowFSM}, a novel agentic framework that leverages Large Language Models (LLMs) combined with prompt chaining and chain-of-thought reasoning to extract accurate FSMs from raw RFC documents. \textit{FlowFSM} systematically processes protocol specifications, identifies state transitions, and constructs structured rule-books by chaining agent outputs. Experimental evaluation across FTP and RTSP protocols demonstrates that \textit{FlowFSM} achieves high extraction precision while minimizing hallucinated transitions, showing promising results. Our findings highlight the potential of agent-based LLM systems in the advancement of protocol analysis and FSM inference for cybersecurity and reverse engineering applications.

\end{abstract}

\vspace{0.1cm}


\section{Introduction}\label{intro}
Finite-State Machines (FSMs) play a critical role in understanding, verifying, and securing network protocols. They model the different states of a protocol and the transitions triggered by various events or inputs, serving as a fundamental basis for protocol verification \cite{fuzzing-book, verified}, vulnerability analysis, and protocol fuzzing \cite{NPF-Survey}. Accurate FSM extraction is critical since vulnerabilities in network services are often deeply tied to unexpected state transitions and misimplementations of protocols \cite{LLM-NPF-Survey}. FSM models have been widely applied, from software testing and security analysis to reverse engineering communication protocols \cite{survey-protocol-reverse-engineering}. Most protocol implementations are guided by the specifications described in the RFC documents \cite{RFCs}, which detail the functional and stateful behavior of network protocols.

Traditional techniques for FSM extraction include static analysis, which infers state transitions by analyzing protocol specifications and program structure without execution \cite{statelifter}, and also dynamic analysis, which observes runtime behaviors \cite{ferry}. Although static analysis offers insights without execution dependencies, it often suffers from scalability issues such as path explosion. In contrast, dynamic methods can miss critical transitions if testing coverage is inadequate \cite{state-selection-algos-imp}.

Recently, large language models (LLMs), originating from transformer-based architectures \cite{attention-is-all-you-need}, have demonstrated significant promise in program analysis tasks such as code completion, repair, and vulnerability detection. They have also shown remarkable proficiency in solving complex computational tasks by generating syntactically correct code \cite{eval-llms-trained-on-code, lmsCanSolve, GenAIForCybersecurity}. Their capabilities extend to FSM extraction and inference \cite{protocolgpt, prosper}, and even protocol fuzzing, where they have been used to guide graybox fuzzing process for improved security analysis \cite{chatafl}. Other FSM tasks, including zero-shot prompting for multi-hop question answering \cite{fsmMHQA}, and aiding developers with robotic FSM modification \cite{fsmRobotic}, were achieved with remarkable performance by LLMs.

We introduce a novel framework, named \textit{FlowFSM}, that combines chain-of-thought reasoning capabilities of LLMs \cite{react},  with a structured prompt chaining strategy for precise extraction of FSMs. Inspired by the principles of \cite{promptChaining}, our approach decomposes the complex task of FSM extraction into a sequence of modular, interpretable steps, where each prompt builds upon the output of the previous one. This chaining strategy not only aggregates incremental gains across steps but also enhances transparency in such LLM-based systems. 

The main contributions of this work can be summarized as follows:
\begin{itemize}
    \item We propose a novel framework, named \textit{FlowFSM}, an AI agent-based flow for FSM extraction using prompt chaining and chain-of-thought techniques.
    
    \item Our method systematically extracts protocol states, message types, and the transitions from specifications of different network protocols.
    
    \item We demonstrate through experimental evaluation that \textit{FlowFSM} significantly improves key metrics such as increased true positives (TPs), reduced false postives (FPs), and balanced F1-scores.
\end{itemize}

The paper is organized as follows: section \ref{rw} outlines the related work. Section \ref{bg} provides background on FSM extraction techniques and LLMs. Section \ref{methodology} details our methodology. Section \ref{evaluation} discusses the research questions and the evaluation of the framework, while section \ref{results} presents the experimental results. Finally, section \ref{conc} concludes the paper and suggests potential directions for future work.

\section{Related Work}\label{rw}
Recent work has explored automating protocol FSM extraction to support security analysis. PROSPER \cite{prosper} presents a framework that leverages LLMs to extract FSMs from RFC documents by combining textual analysis with artifact mining. It uses a multi-step pipeline involving artifact extraction, prompt engineering, and few-shot learning to identify states, transitions, and events. Evaluation across 30 RFCs showed that PROSPER outperforms rule-based baselines, achieving 1.3× more true positives and 6.5× fewer false positives in FSM recovery. ProtocolGPT \cite{protocolgpt} introduces a system for extracting FSMs directly from protocol implementations using OpenAI's GPT-4. It employs LLMs to first filter and retrieve FSM-relevant code regions and then infer states and transitions through sequential prompt queries, embedding retrieval, and structured output generation. Experiments on six protocol implementations achieved over 90\% precision and recall, and integration with fuzzers such as AFLNet \cite{aflnet} yielded up to 10\% more code coverage compared to FSMs extracted from specifications. RFCNLP \cite{rfcnlp} proposes a hybrid machine learning and rule-based approach to extract FSMs from RFC documents using technical language embeddings, zero-shot semantic parsing, and heuristics to generate intermediate representations. The extracted FSMs were successfully used to synthesize attacks on TCP and DCCP, and the evaluation showed that even partial FSMs enable attacker synthesis, demonstrating the framework’s robustness despite incomplete extraction due to natural language ambiguities. Hermes \cite{hermes} proposes a neural and symbolic framework for synthesizing FSMs from cellular network specifications using a neural parser and logical IR synthesis, achieving up to 87\% extraction accuracy and allowing discovery of new vulnerabilities and deviations in commercial 4G/5G implementations.

\begin{figure}[h]
    \centering
    \hspace*{-1cm}
    \includegraphics[scale=0.55]{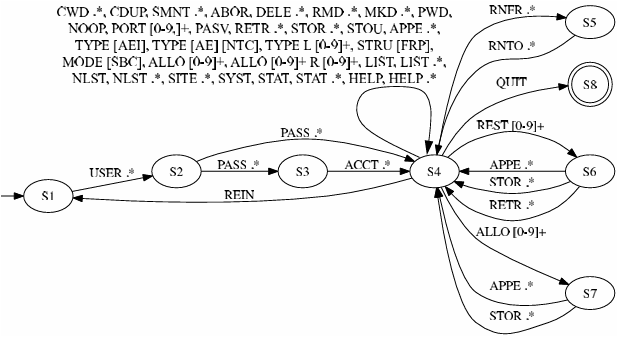}
    \caption{FTP protocol's finite-state machine as defined in RFC-959.}
    \label{fig:ftp_fsm}
\end{figure}

\section{Background}\label{bg}
This section explains background on the protocol's finite-state machine and LLMs. First, we discuss the pros and cons of the current existing approaches for FSM extraction. We then discuss LLM capabilities and some of their applications in the cybersecurity field.

\subsection{Finite-State Machine}
FSMs provide a formal model to represent the operational logic of network and communication protocols. An FSM defines a finite set of protocol states, the inputs triggering transitions, and the conditions that govern these transitions. During protocol interaction, specific events or messages lead to state changes, enabling structured and predictable communication. Inputs are typically the commands or responses exchanged between clients and servers, formatted according to protocol specifications to ensure interoperability.
The File Transfer Protocol (FTP) exemplifies the use of FSMs in practice, modeling states such as Not Connected, Authorization, Transaction, and Update. As shown in Figure \ref{fig:ftp_fsm}, transitions between these states are triggered by commands like CONNECT, USER, PASS, PORT, and QUIT, facilitating operations from authentication to file transfer. FSM inference in protocols like FTP is vital for understanding behavior, verifying implementations, and detecting deviations that may lead to security vulnerabilities.
Various FSM extraction techniques exist, including network traffic analysis \cite{netplier}, static code analysis \cite{statelifter}, dynamic program analysis \cite{ferry}, and natural language processing approaches \cite{rfcnlp}. Each offers unique strengths and limitations. Network traffic analysis can reverse-engineer FSMs by observing real-world exchanges but may miss rare or unexercised transitions. Static analysis offers deep insights without execution but often struggles with scalability and complex code paths. Dynamic analysis captures real-time behavior but heavily depends on input quality and coverage. NLP-based approaches extract FSMs from specifications but are limited by ambiguities and inconsistencies inherent in natural language documents.

\subsection{Large Language Models}
LLMs have become strong tools in artificial intelligence, capable of performing lots of tasks including text generation, classification, and reasoning \cite{deepseek}. Recent research has expanded their application to the domains of cybersecurity and protocol analysis. LLMs have been used for tasks such as guiding protocol fuzzing \cite{chatafl}, phishing site detection \cite{chatphishdetector}, and automated penetration testing through agent collaboration \cite{pentestagent}. In software engineering, LLMs have demonstrated strong abilities in automation \cite{erpa, lmrpa, lmvrpa, llm-daas}, test generation, and escaping coverage plateaus in fuzzing \cite{codamosa}, as well as augmenting graybox fuzzing through intelligent input generation \cite{chatfuzz}. Further advancements have shown that LLM-driven seed generation significantly improves fuzzing efficiency compared to traditional methods \cite{seedmind}. Retrieval-augmented generation (RAG) \cite{rag} combined with chain-of-thought reasoning techniques has been proven to improve the accuracy and reliability of LLMs in network packet generation \cite{ehna}. Despite these successes, challenges remain, including randomness in model outputs, context window limitations, and difficulties handling complex static analysis tasks, necessitating incremental prompting and human guidance to achieve high precision in security-critical applications.

\section{Methodology}\label{methodology}
This section presents the methodology of our work, that proposes an agentic flow named \textit{FlowFSM}, an innovative approach for FSM extraction from raw RFC documents. The extracted FSM contains defined states within the network protocol, commands that accommodate each state, and transitions in the states. The methodology can be divided into two major steps, which are: RFC documents processing, and FSM extraction using prompt chaining.

\begin{figure}[h]
    \centering
    \hspace{-1.227cm}
    \includegraphics[width=10cm]{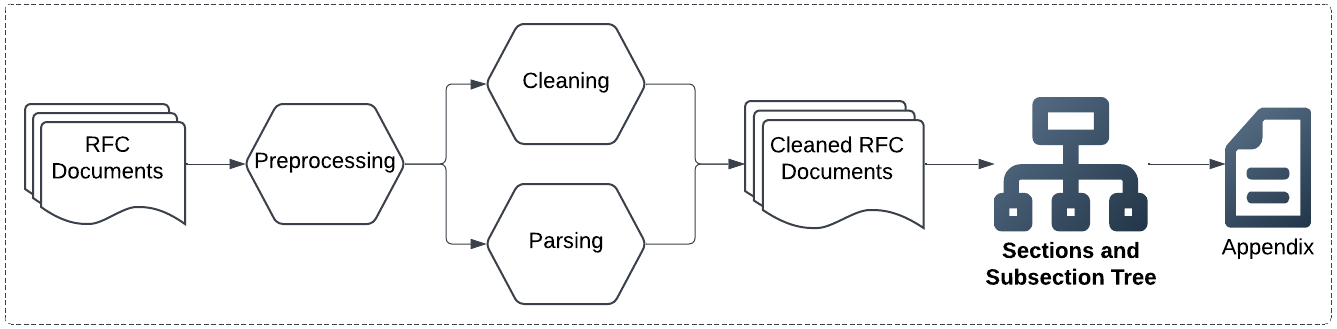}
    \caption{RFC Documents Processing Pipeline}
    \label{fig:rfc-pipeline}
\end{figure}

\subsection{RFC Document Processing}
Parsing RFC documents is crucial for extracting protocol specifications and FSM structures. RFCs are first preprocessed to remove page headers, footers, and formatting artifacts, producing continuous raw text \( T \in \mathbb{S} \), where \(\mathbb{S}\) denotes the set of all string sequences.
We define an RFC document as a hierarchical tree:

\[
 \mathcal{T} = (N, E) 
\]

where \( N \) is the set of nodes (sections) and \( E \subseteq N \times N \) is the set of parent-child edges. Each node \( n \in N \) is represented as a structured \texttt{JSON} object containing a title, body, path, and a list of subsections. A parsing function \( P: \mathbb{S} \to \mathcal{T} \) recursively segments \( T \) based on section numbering patterns, such that:

\[
P(T) = \bigcup_{i=1}^{k} P(S_i)
\]

where \( S_i \) are contiguous subsections identified by regular expressions. Leaf nodes are defined as nodes \( n \in N \) where:

\[
\forall m \in N, (n, m) \notin E
\]

All leaf node bodies are collected into a set of chunks for downstream processing. The final appendix structure \( A \) is generated as an ordered listing of all paths in \( \mathcal{T} \), enhancing LLM reasoning for FSM extraction and protocol command identification. A summary of the whole RFC processing pipeline can be represented in Figure \ref{fig:rfc-pipeline}.

\begin{figure}[h]
    \centering
    \hspace{0cm}
    \includegraphics[scale=0.85]{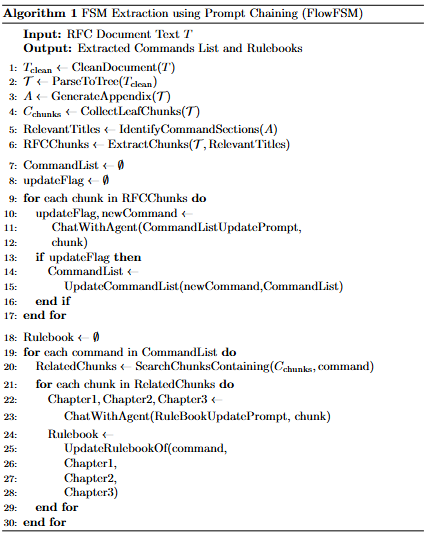}
    \caption{Pseudocode of the \textit{FlowFSM} algorithm}
    \label{fig:pseudocode}
\end{figure}


\subsection{FSM Extraction using Prompt Chaining}
Extracting FSMs from protocol specifications requires navigating complex RFC document structures and accurately interpreting state transition semantics. Our approach combines prompt chaining with LLM reasoning to systematically construct a structured rulebook, which serves as the formal representation of the protocol's FSM.

\begin{figure}[h!]
    \centering
    \raisebox{0pt}[\height][0pt]{\hspace*{-0.4cm}%
        \includegraphics[scale=0.38]{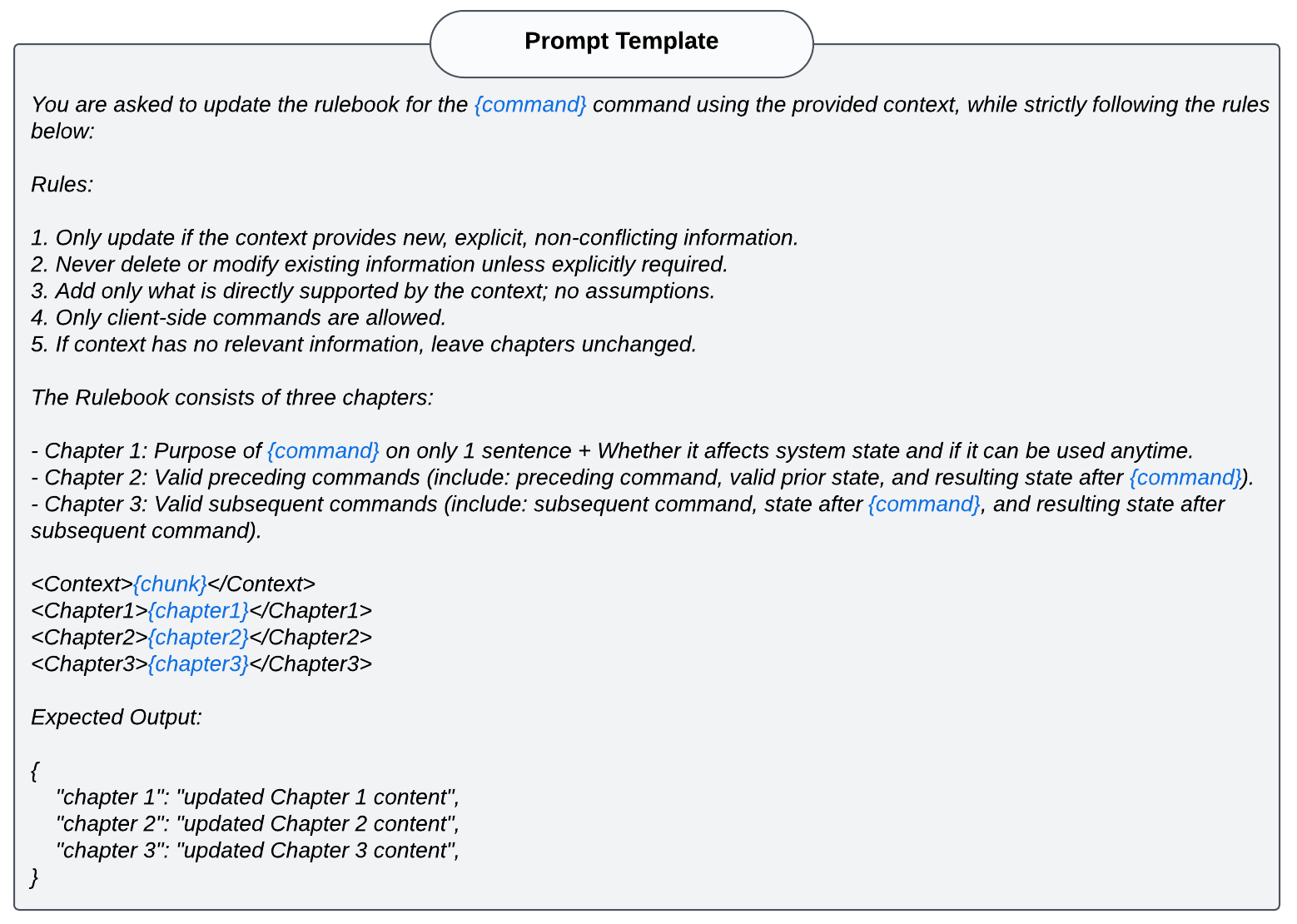}%
    }
    \caption{Three-stage prompt chaining template for iterative rulebook update}
    \label{fig:prompt_template}
\end{figure}

Prompt chaining involves breaking down the extraction process into a sequence of guided interactions. At each stage, a prompt is carefully constructed to refine the output progressively, starting from identifying relevant sections, isolating candidate commands, and synthesizing structured state-transition information. 


Formally, let \( \mathcal{P}_i \) denote the \( i \)-th prompt and \( R_i \) the corresponding model response. The overall FSM extraction process can be formulated as:

\[
R_{i+1} = \mathcal{M}(\mathcal{P}_i(R_i))
\]

where \( \mathcal{M} \) represents the LLM model, and each \( \mathcal{P}_i \) depends on the output of the previous step. This iterative refinement minimizes hallucinations, enforces consistency, and improves the robustness of the extraction. By chaining prompts in a controlled flow, we decompose the complex FSM extraction task into manageable subtasks, significantly enhancing both extraction accuracy and reliability of results. A pseudocode of the algorithm can be shown in Figure \ref{fig:pseudocode}.

\vspace{0.2cm}

\subsection{Prompt Chaining Architecture}
As shown in the prompt template in Figure \ref{fig:prompt_template}, we implement a three-stage chaining process:

\begin{enumerate}
    \item \textbf{Command Extraction}: extracts candidate commands from RFC chunks through text analysis. It classifies these commands into functional categories. This phase produces an initial command inventory annotated with basic operational descriptions.

\item \textbf{State Transition Analysis}: identifies precondition and postcondition states for each extracted command. It maps all allowable command sequences within the protocol specification.

\item \textbf{Rulebook Synthesis}: formalizes the extracted information into the three-component rulebook structure. A sample rulebook example is shown in Figure \ref{box:pass_rulebook}.
\end{enumerate}


\subsection{Rulebook Structure}
The rulebook organizes each protocol command through three fundamental components that collectively define the command's behavior within the state machine. The rulebook organizes protocol commands into three main chapters: 

\begin{enumerate}
    \item \textbf{Command Purpose \& Outlines:} establishes the command's functional role within the protocol. This section precisely documents the command's effects on system state, including any state transitions it triggers, and identifies all temporal constraints governing when the command can be legally executed within the protocol sequence.

    \item \textbf{Valid Preceding Commands:} defines the precondition states necessary for command execution. This includes specifying any mandatory prior commands that must precede the current command. Beyond simple sequencing, this section validates permissible command chains through comprehensive state dependency analysis, ensuring only valid protocol flows are permitted.

    \item \textbf{Valid Subsequent Commands:} outlines all possible legal state transitions following command execution. For each permitted subsequent command, this section documents the precise post-condition states that result from the transition while explicitly identifying and prohibiting invalid state transitions that could violate protocol specifications.
\end{enumerate}

\vspace{0.3cm}

\begin{figure}[t]
  
{\small
\begin{minipage}{\linewidth}
    \hspace*{2.2em}
\begin{tcolorbox}[colframe=black, colback=white, title=Sample Rulebook Output for \texttt{PASS} Command, fonttitle=\bfseries, label={box:pass_rulebook}]
\textbf{Chapter 1: Command Purpose \& Outlines}

The \texttt{PASS} command is used to send the password, completing the user's identification process. Executing \texttt{PASS} transitions the system to a state where account information may be required. It cannot be used at any time, it must immediately follow the \texttt{USER} command.

\vspace{0.3cm}

\textbf{Chapter 2: Valid Direct Preceding Commands/Methods}

- \textbf{Preceding Command:} \texttt{USER} \\
- \textbf{System State:} The system must have received the \texttt{USER} command. \\ 
-  \textbf{Changes System State:} Yes, transitions to a state where account information may be required.

\vspace{0.3cm}

\textbf{Chapter 3: Valid Direct Subsequent Commands/Methods}

- \textbf{Subsequent Command:} \texttt{RETR}  \\
-  \quad\textbf{System State:} User must be logged in. \\  
-  \quad\textbf{Changes System State:} Yes, retrieves a file.

- \textbf{Subsequent Command:} \texttt{TYPE}  \\
-  \quad\textbf{System State:} User must be logged in. \\  
-  \quad\textbf{Changes System State:} Yes, sets the file transfer type.
\end{tcolorbox}
\end{minipage}
}
\begin{center}
\small\textbf{Figure \ref{box:pass_rulebook}:} Sample generated rule-book output for the FTP \texttt{PASS} command showing its purpose, valid preceding, and subsequent commands.
\end{center}
\end{figure}

\subsection{Implementation}
We have implemented our framework, named \textit{FlowFSM}, on top of CrewAI \cite{crewai}, an open-source Python multi-agent-based library. CrewAI is a framework independent of LangChain and other AI-agent-based frameworks. It offers a modular 'Crew' abstraction to define autonomous LLM agents and flows for event-driven orchestration, allowing precise task sequencing. CrewAI integrates with multiple LLM providers such as OpenAI, Anthropic, Azure, and others, giving flexibility in model selection and seamless scaling. It's built-in tools and high-level APIs simplify the integration of utilities such as web search, file parsing, and vector store retrieval for RAG applications. Within this system, \textit{FlowFSM} packages RFC processing, context retrieval, and stepwise FSM extraction tasks into separate modules. We open source \textit{FlowFSM}'s source code at: \url{https://github.com/YoussefMaklad/FlowFSM}.

\section{Experimental Design and Evaluation}\label{evaluation}
We evaluate the framework by evaluating \textit{FlowFSM}'s ability to extract protocol state machines. This evaluation of \textit{FlowFSM} aims to answer the following research questions:
\begin{itemize}
    \item \textbf{RQ1:} How accurate is \textit{FlowFSM} in extracting protocol state machines? 
    \item \textbf{RQ2:} Does \textit{FlowFSM} generalize to different protocol specifications?
\end{itemize}

We conducted experiments using a combination of three powerful LLMs: \textit{llama3.3-70b-versatile}, \textit{deepseek-r1-distill-llama-70b}, and \textit{llama3-70b-8192}. These models were selected based on their performance and large context windows. The experiments were performed on two different network protocols: the File Transfer Protocol (FTP) and the Real-Time Streaming Protocol (RTSP). These protocols were chosen to represent diverse characteristics in protocol structure, interaction complexity, and document style.

\subsection{Evaluation Metrics}
To quantify the performance of \textit{FlowFSM}, we use the following metrics:
\begin{itemize}
    \item \textbf{True Positives (TP):} The number of extracted state transitions that were manually verified to be correct according to the official protocol behavior.
    \item \textbf{False Positives (FP):} The number of extracted transitions that were incorrect or hallucinated, deviating from the actual protocol specifications.
    \item \textbf{False Negatives (FN):} The number of valid state transitions present in the protocol but missed by the extraction process.
    \item \textbf{Precision:} The ratio of correctly extracted transitions to all extracted transitions (higher values indicate fewer false positives).
    \item \textbf{Recall:} The ratio of correctly extracted transitions to all valid transitions in the protocol (higher values indicate fewer false negatives).
    \item \textbf{F1-Score:} The harmonic mean of precision and recall, providing a balanced measure of extraction accuracy
\end{itemize}

The correctness of extracted transitions was assessed through manual validation by cross-referencing with the protocol RFC documents and standard references. The balance between true positives and false positives provides insight into the reliability of \textit{FlowFSM} in FSM extraction across different protocols. The evaluation metrics can be calculated as follows:

\[
\text{Precision} = \frac{\text{TP}}{\text{TP} + \text{FP}} \times 100 \qquad 
\text{Recall} = \frac{\text{TP}}{\text{TP} + \text{FN}} \times 100
\]
\[
\text{F1-Score} = 2 \times \frac{\text{Precision} \times \text{Recall}}{\text{Precision} + \text{Recall}} \times 100
\]


\section{Experimental Results and Discussion}\label{results}
In this section, we present the experimental results of evaluating \textit{FlowFSM} on the FTP and RTSP protocols. We assess the extraction performance based on the number of true positives (TP) and false positives (FP) identified in the inferred-state machines. Table~\ref{tab:extraction-results} summarizes the results obtained in the two protocols. The metrics reflect the robustness of \textit{FlowFSM} when processing different protocol transitions and interaction complexities.

\begin{table}[t]
\hspace*{-0.25cm}
\centering
\scalebox{1.3}{%
\begin{tabular}{|c|c|c|c|c|c|c|}
\hline
\textbf{Protocol} & \textbf{TP} & \textbf{FP} & \textbf{FN} & \textbf{Precision} & \textbf{Recall} & \textbf{F1-Score} \\ \hline
FTP  & 90 & 18 & 12 & 83.33\% & 88.24\% & 85.71\% \\ \hline
RTSP & 18 & 4  & 3  & 81.82\% & 85.71\% & 83.72\% \\ \hline
\end{tabular}
}
\vspace{0.15cm}
\caption{FSM Extraction Results for FTP and RTSP Protocols}
\label{tab:extraction-results}
\end{table}


The experimental results demonstrate \textit{FlowFSM}'s effectiveness in protocol FSM extraction, with both protocols achieving precision scores above 81\% and recall above 85\%. For FTP, the system correctly identified 90 true state transitions (TP) while generating only 18 false positives (FP), yielding a precision of 83.33\%. The 12 false negatives (FN) represent cases where valid transitions were missed, resulting in 88.24\% recall and an F1-score of 85.71\%. RTSP extraction showed comparable performance with 81.82\% precision and 85.71\% recall and 83.72\% for F1-score. Three key technical observations emerge from these results:

\begin{enumerate}
    \item \textbf{Recall-Precision Tradeoff (Answering RQ1)}: The higher recall scores (88.24\%, 85.71\%) compared to precision indicate \textit{FlowFSM}'s conservative approach as it prioritizes capturing more potential transitions at the cost of requiring subsequent validation. The high F1-scores further validate this balance, showing that the tradeoff does not compromise overall accuracy. This proves advantageous for security applications where missing transitions (false negatives) could be more critical than extra candidates.
    
    \item \textbf{Consistent Performance Across Protocols (Answering RQ2)}: The marginal difference between FTP and RTSP precisions (83.33\% and 81.82\% respectively) suggests the method generalizes well despite RTSP's more complex session management requirements, and F1-scores, differing by only 1.99\%, reinforce this consistency. The 1.51\% precision variance falls within expected bounds for different protocol types.
\end{enumerate}

The results particularly highlight \textit{FlowFSM}'s robustness against two common LLM failure modes: (1) hallucination of non-existent transitions (controlled via precision metrics) and (2) omission of valid protocol behaviors (measured through recall). The F1-scores confirm that the framework maintains this robustness holistically. The balanced performance of \textit{FlowFSM} suggests the prompt chaining strategy effectively constrains LLM outputs while maintaining comprehensive awareness of protocol specifications.

\section{Conclusion and Future Work}\label{conc}
Finite-State Machines (FSMs) are fundamental for understanding, verifying, and securing network protocols, providing structured models of protocol behavior, and guiding vulnerability discovery. Traditional FSM extraction methods, such as static or dynamic analysis, face significant challenges, including path explosion and incomplete transition coverage. Recent improvements in LLMs have proven strong capabilities in program analysis, code generation, and FSM-related tasks, offering new opportunities for protocol analysis. In this paper, we presented \textit{FlowFSM}, a novel agent-based framework which uses prompt chaining and chain-of-thoughts to systematically extract FSMs from protocol specifications and construct a structured rule-book. Our approach identifies protocol states, messages, and transitions with promising results. Experimental evaluation demonstrated that \textit{FlowFSM} significantly enhances FSM extraction quality, achieving higher true positive rates while reducing false positives. Future work will explore extending \textit{FlowFSM} to more diverse protocol families and further optimizing multi-agent collaboration strategies to enhance scalability and robustness, as a limitation of \textit{FlowFSM} it it's high computational and runtime cost. We also plan to integrate \textit{FlowFSM} in protocol fuzzing frameworks to guide the fuzzing process.


\bibliographystyle{IEEEtran}
\bibliography{references}

\end{document}